\documentclass[10pt,twocolumn,letterpaper]{article}

\usepackage{iccv}
\usepackage{times}
\usepackage{epsfig}
\usepackage{graphicx}
\usepackage{amsmath}
\usepackage{amssymb}

\usepackage{booktabs} 
\usepackage{makecell}
\usepackage{multirow}
\usepackage{arydshln}
\usepackage{overpic}
\usepackage{diagbox}

\usepackage{adjustbox}
\usepackage{array}

\usepackage[pagebackref=true,breaklinks=true,letterpaper=true,colorlinks,bookmarks=false]{hyperref}

\iccvfinalcopy 

\ificcvfinal\fi

\begin{document}

\title{Learning to Resize Images for Computer Vision Tasks}

\author{Hossein Talebi and Peyman Milanfar \\
Google Research\\
}

\maketitle
\ificcvfinal\thispagestyle{empty}\fi

\begin{abstract}
For all the ways convolutional neural nets have revolutionized computer vision in recent years, one important aspect has received surprisingly little attention: the effect of image size on the accuracy of tasks being trained for. Typically, to be efficient, the input images are resized to a relatively small spatial resolution (e.g. $224\times224$), and both training and inference are carried out at this resolution. The actual mechanism for this re-scaling has been an afterthought: Namely, off-the-shelf image resizers such as bilinear and bicubic are commonly used in most machine learning software frameworks. But do these resizers limit the on-task performance of the trained networks? The answer is yes. Indeed, we show that the typical linear resizer can be replaced with learned resizers that can substantially improve performance. Importantly, while the classical resizers typically result in better perceptual quality of the downscaled images, our proposed learned resizers do not necessarily give better visual quality, but instead improve task performance.

Our learned image resizer is jointly trained with a baseline vision model. This learned CNN-based resizer creates machine friendly visual manipulations that lead to a consistent improvement of the end task metric over the baseline model. Specifically, here we focus on the classification task with the ImageNet dataset~\cite{russakovsky2015imagenet}, and experiment with four different models to learn resizers adapted to each model. Moreover, we show that the proposed resizer can also be useful for fine-tuning the classification baselines for other vision tasks. To this end, we experiment with three different baselines to develop image quality assessment (IQA) models on the AVA dataset~\cite{murray2012ava}.
\end{abstract}

\section{Introduction}

The emergence of deep neural networks along with large scale image datasets has led to major breakthroughs in machine visual recognition. Images in such datasets are typically obtained from the web and as a result have gone through various capture pipelines and post-processing steps. Besides these generally unknown processing operations, efficient training of visual recognition CNNs requires additional image augmentations such as spatial resizing.

\begin{figure}[t]
\vspace{-0 mm}
\begin{center}
\includegraphics*[viewport=1 1 260 150,scale=0.9]{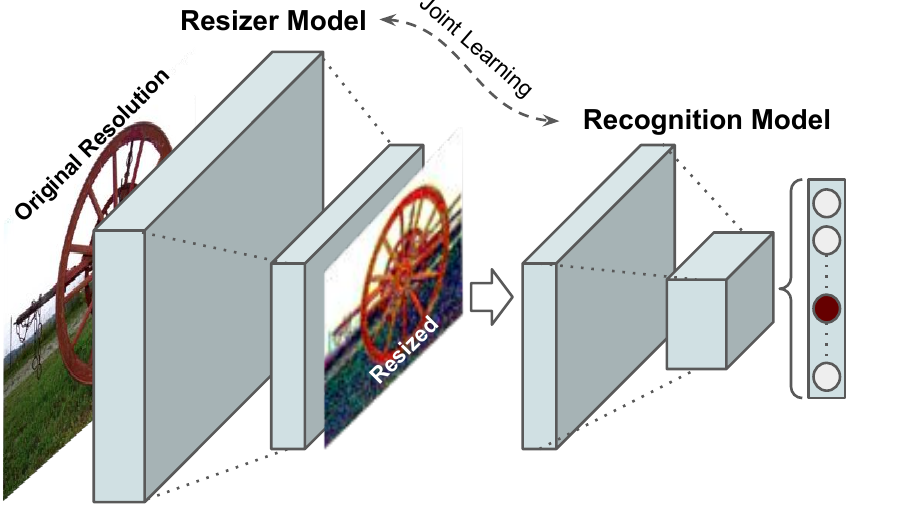}
\end{center}
\vspace{-4 mm}
{\caption{Our proposed framework for joint learning of the image resizer and recognition models.} \label{fig:diagram}}
\vspace{-4 mm}
\end{figure}

\begin{table}[t]
\renewcommand{\arraystretch}{1.0}
\footnotesize
    \centering
    \begin{tabular}{lcccc} \toprule
        \multirow{2}{*} & &  \multicolumn{2}{c}{Top-1 Error $\downarrow$} \\\cline{3-4}
        \makecell{Task} & Model & Bilinear Resizer & Proposed Resizer\\\midrule
        \parbox[c]{0mm}{\multirow{4}{*}{\rotatebox[origin=c]{90}{\scriptsize{Classification}}}}
        & Inception-v2~\cite{szegedy2016rethinking} & 26.7\%  & 24.0\% \\
        & DenseNet-121~\cite{huang2017densely} & 33.1\%  & 29.8\%  \\
        & ResNet-50~\cite{he2016deep} & 24.7\%  & 23.0\%  \\
        & MobileNet-v2~\cite{sandler2018mobilenetv2} & 29.5\%  & 28.4\%  \\ \cmidrule{3-4}
        \multirow{2}{*} & &  \multicolumn{2}{c}{PLCC $\uparrow$} \\\cline{3-4}
        &  & Bicubic Resizer & Proposed Resizer\\\midrule
        \parbox[c]{0mm}{\multirow{3}{*}{\rotatebox[origin=c]{90}{\scriptsize{IQA}}}}
        & Inception-v2~\cite{szegedy2016rethinking} & 0.662  & 0.686 \\
        & DenseNet-121~\cite{huang2017densely} & 0.662  & 0.683  \\
        & EfficientNet-b0~\cite{tan2019efficientnet} & 0.642  & 0.671  \\\midrule
    \end{tabular}
    \vspace{-1mm}
    \caption{Summary of our results for image classification on ImageNet~\cite{russakovsky2015imagenet}, and image quality assessment (IQA) on the AVA dataset~\cite{murray2012ava}.}
    \label{tab:summary}
    \vspace{-6 mm}
\end{table}

Image down-scaling is the most commonly used pre-processing module in classification models. The main reasons for spatial resizing are: (1) mini-batch learning through gradient descent requires the same spatial resolution for all images in a batch, (2) memory limitations prohibit training CNNs at high resolutions, and (3) large image sizes lead to slower training and inference. Given a fixed memory budget, there is a trade-off between the memory occupied by the spatial resolution and the batch size. This trade-off can have a significant impact on the accuracy of recognition CNNs~\cite{sabottke2020effect, touvron2019fixing,huang2019gpipe,karras2017progressive}.

Currently rudimentary resizing methods such as nearest neighbor, bilinear, and bicubic are among the top adopted image resizers visual recognition systems. These resizers are fast, and can be flexibly integrated into the train and test frameworks. However, these methods were developed decades before deep learning became the mainstream solution for visual recognition tasks, and hence are not optimized or adequate for machine perception.

Recent research on recognition-aware image processing has shown promising results on improving accuracy of classification models and simultaneously preserving the perceptual quality~\cite{liu2019transferable, scheirer2020bridging}. This class of methods keep the classification model fixed, and only train the  enhancement module. Meanwhile, there has been some effort on joint learning of both the pre-processor and the recognition model~\cite{bai2018finding,zhang2018super,liu2018disentangling,sharma2018classification,liu2017image,li2018end,zhao2020thumbnet}. These algorithms set up training frameworks with hybrid losses that allow for learning better enhancement and recognition, concurrently. In practice, however, a recognition pre-processing operation such as resizing should not be optimized for better perceptual quality, because the end goal is for the recognition network to produce accurate results, not for the intermediate image to ``look good" to a human observer.

In this paper we propose a novel image resizer that is jointly trained with classification models (see Figure~\ref{fig:diagram}), and is specifically designed to improve classification performance (see Table~\ref{tab:summary}). To summarize our contributions:

\begin{itemize}

	\item We couple our resizer with various classification models and show that it effectively adapts to each model and consistently improves over the baseline image classifier.
	
	\item The proposed resizer is not constrained by any pixel or perceptual loss, therefore our results present machine adaptive visual effects that differ from conventional image processing and super-resolution results.
	
	\item The proposed resizer model allows for down-scaling images at arbitrary scaling factors, hence we can conveniently search for the most optimal resolution for an underlying task.
	
	\item We expand the application of the proposed resizer to image quality assessment (IQA) and show that it successfully adapts to this task.
	
\end{itemize}

We purport the proposed method is the first pre-processing model that is specifically developed for vision tasks, and that aims to replace off-the-shelf resizers. Perhaps remote machine learning inference can be mentioned as a distinguishing application of the proposed resizer. In remote inference, transferring full resolution Mega pixel images from clients to servers imposes a significant latency on the system. While down-scaling by off-the-shelf resizers on the client side may reduce the latency issue, it may also negatively impact the recognition performance. The proposed resizer, therefore, can be an alternative to the off-the-shelf resizers to effectively reduce the expected drop in the recognition performance.

\noindent Next, we briefly survey the research works related to this paper. Then, in Section~\ref{sec:proposed} the proposed resizer model is discussed in detail. In Section~\ref{sec:experiments} our results are presented, and finally we conclude in Section~\ref{sec:conclusion}.

\begin{figure}
    \centering
    \scriptsize

    \vspace{-2mm}
    \begin{overpic}[width=.49\linewidth]{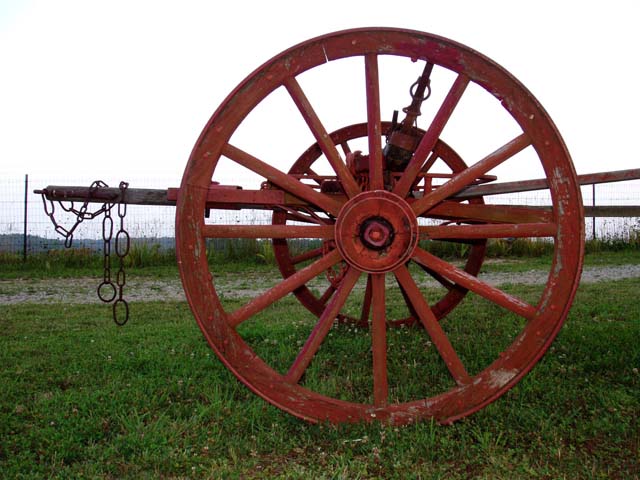}
    \put(2,70){\begin{color}{black}Original \end{color}}
    \put(2,64){\begin{color}{black} ($480\times640$)\end{color}}
    \end{overpic}
    %
    %
    \begin{overpic}[width=.49\linewidth]{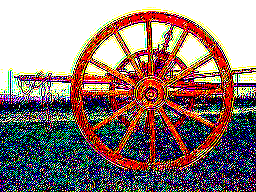}
    \put(2,70){\begin{color}{black} Learned Resizer \end{color}}
    \put(2,64){\begin{color}{black} ($192\times256$)\end{color}}
    \end{overpic}
    %
    %
    \begin{overpic}[width=.244\linewidth]{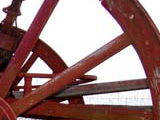}
    \end{overpic}
    \begin{overpic}[width=.244\linewidth]{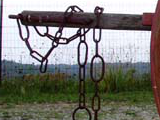}
    \end{overpic}
    %
    %
    %
    \begin{overpic}[width=.244\linewidth]{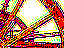}
    \end{overpic}
    \begin{overpic}[width=.244\linewidth]{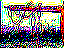}
    \end{overpic}
    %
    %
    \vspace{-3mm}
    
    \caption{Example of the proposed resizer trained for image classification on the ImageNet dataset~\cite{russakovsky2015imagenet}. The baseline classification model is Inception-v2~\cite{szegedy2016rethinking}, and it is jointly trained with the resizer model shown in Fig.~\ref{fig:cnn}. The resized image fits the classification task better than the existing pre-processing resizers such as bilinear and bicubic.}
    \label{fig:comp_resized}
    \vspace{-6 mm}
\end{figure}

\section{Related Work}

\begin{figure*}[!t]
\vspace{-3 mm}
\begin{center}
\includegraphics*[viewport=1 1 720 165,scale=0.64]{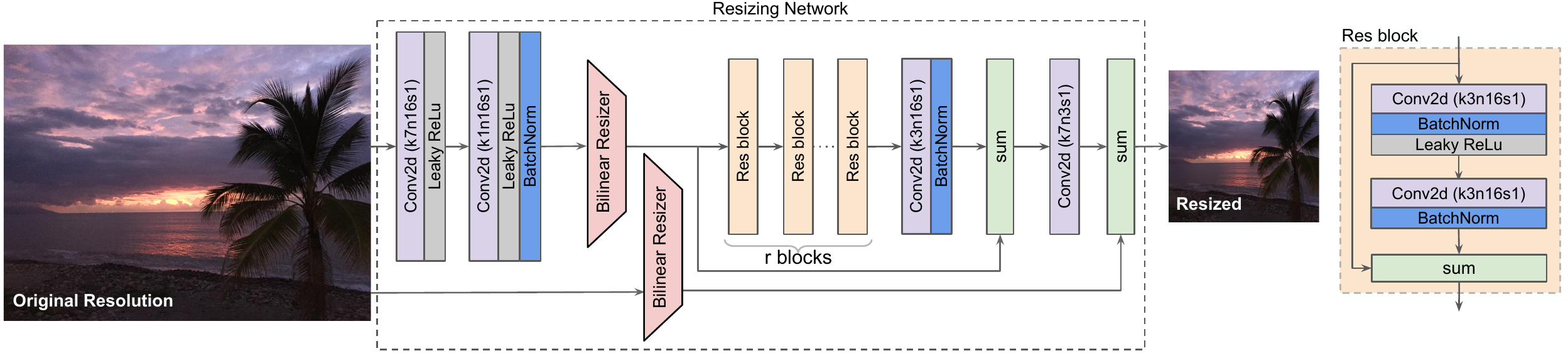}
\end{center}
\vspace{-5 mm}
{\caption{Our proposed CNN model for resizing images. The bilinear feature resizer allows for arbitrary up/down-scaling factors.} \label{fig:cnn}}
\vspace{-2 mm}
\end{figure*}

There is ample reason to believe, given the past literature, that an optimized pre-processing module can improve performance of computer vision systems. For example, concurrent optimization of object recognition and enhancement tasks dates back to Zeiler et al. \cite{zeiler2010deconvolutional}, where they used deconvolutional networks to learn robust features for image synthesis and analysis. Namboodiri et al.~\cite{namboodiri2011systematic} show that enhancement algorithms such as super-resolution should be evaluated by classification driven metrics. More recently, a super-resolution algorithm~\cite{haris2018task} trained using an object detection loss showed superior results compared to the conventional super-resolution algorithms. Gondal et al.~\cite{waleed2018unreasonable} studied the impact of super-resolution on recognition tasks by super-resolving down-sampled ImageNet~\cite{russakovsky2015imagenet} images and comparing classification accuracy for various up-scaling methods. They concluded that the choice of resizing method has a significant impact on the performance. More recently, Singh et al.~\cite{singh2019dual} introduced a method for up-scaling very small images to improve face and digit recognition.

The impact of compression on recognition has also been studied in~\cite{luo2020rate,suzuki2019image,tahboub2017quality}. Luo et al.~\cite{luo2020rate} show that JPEG quantization coefficients can be optimized to obtain lower bit-rates and at the same time preserve perceptual quality and recognition performance. Recently, several pre-editing methods for more efficient compression without sacrifice of the classification accuracy have been introduced~\cite{suzuki2019image,talebi2020better}. Typically these methods rely on some kind of rate-distortion-accuracy optimization.

Sharma et al. \cite{sharma2018classification} train a generic enhancement model through joint learning with the classification network. This enhancement model does not alter the spatial resolution, and is trained via an $L_2$ loss on the enhanced image added to a cross-entropy classification loss. Denosing is yet another enhancement operation that has been successfully employed to improve the perceptual quality and recognition accuracy~\cite{diamond2017dirty,yim2017enhancing}. The approach of Diamond et al.~\cite{diamond2017dirty} focuses on low light imaging scenarios and shows that the best image processing algorithms for computer vision tasks are different from existing methods developed to produce visually pleasing photos. Also, the algorithm of Li et al.~\cite{li2017aod} shows that, dehazing can boost object detection and recognition performance on natural hazy images.

Recently Liu et al.~\cite{liu2019transferable} proposed an approach to improve machine interpretability of images by optimizing the recognition loss on the image processing network. They study super-resolution, denoising, and JPEG-deblocking as pre-processing operations, and show that the recognition performance gain can transfer when evaluated on different architectures and tasks. Our approach differs with \cite{liu2019transferable} in that (1) we exclusively focus on improving the recognition performance, regardless of the perceptual quality, and (2) our pre-processing resizer is jointly trained with the recognition model, and consequently better adapts to the recognition architecture. This means the proposed model is not constrained to learn specific enhancements (e.g. denoising or deblurring), but rather it freely learns some unique machine friendly effects that result in a recognition gain (see Figure~\ref{fig:comp_resized}). This characteristic suits our model for applications where visually pleasing images are not the end goal.

\section{Proposed Framework}
\label{sec:proposed}

In this section we introduce our resizer model, and discuss how we deploy it for training and testing image classification and IQA models.

Our resizer model is designed to be easily trainable, so it can be plugged into various learning frameworks and tasks. Also, it handles any arbitrary scaling factor, including up and down-scaling. This allows us to explore the resolution vs batch size trade-off, and as a result find the optimal resolution for the task in hand. In terms of performance, ideally, the net gain obtained by such adaptive resizing should surpass the extra computational complexity that the resizer adds to the system. These constraints make it almost impossible to use the existing super-resolution models~\cite{lim2017enhanced,ledig2017photo,wang2018esrgan,zhou2019kernel,cai2019toward,milanfar2017super,berthelot2020creating}. On the other hand, image re-scaling methods such as bilinear and bicubic are, on their own, not trainable, and hence not suitable for this task. To this end, we design a model that satisfies these criteria.

\subsection{Resizer Model}

\begin{table}[t]
\vspace{-2 mm}
\renewcommand{\arraystretch}{1.0}
\footnotesize
    \centering
    \begin{tabular}{lcccc} \toprule
        \diagbox[width=7em]{Filters}{Blocks} & $r=1$ & $r=2$ & $r=3$ & $r=4$ \\\cmidrule{1-5}
        n=16 & 11.87 & 16.48  & 21.08 & 25.69   \\
        n=32 & 38.08  & 56.51  & 74.94 & 93.37 \\\midrule
    \end{tabular}
    \vspace{-0.5mm}
    \caption{Number of parameters in the proposed resizer model are given in thousands. The number of residual blocks and the number of convolutional filters in Figure~\ref{fig:cnn} are varied in this table.}
    \label{tab:cnn_params}
    \vspace{-6mm}
\end{table}

Our proposed resizer architecture is shown in Figure~\ref{fig:cnn}. Perhaps the most important characteristics of this model are (1) the bilinear feature resizing, and (2) the skip connection that accommodates combining the bilinearly resized image and the CNN features. The former factor allows for incorporation of features computed at original resolution into the model. Also, the skip connection accommodates for an easier learning process, because the resizer model can directly pass the bilinearly resized image into the baseline task. Note that the bilinear feature resizer shown in Figure~\ref{fig:cnn} acts as a feed-forward bottleneck (down-scaling), but in principle it can also act as an inverse bottleneck as well (up-scaling). It is worth noting that unlike typical encoder-decoder architectures~\cite{mao2016image}, the proposed architecture allows for resizing an image to any target size and aspect ratio. It is also important to highlight that performance of the learned resizer is barely dependent on the bilinear resizer choice, meaning it can be safely replaced with other off-the-shelf methods such as bicubic or Lanczos. 

The residual blocks used in our model are inspired by~\cite{gross2016training,ledig2017photo}. There are $r$ identical residual blocks in our model and in our experiments we set $r=1$ or $2$. All intermediate convolutional layers have $n=16$ kernels of size $3\times3$. The first and the last layers consist of $7\times7$ kernels. The larger kernel size in the first layer allows for a $7\times7$ receptive field on the original image resolution. We also use batch normalization layers~\cite{ioffe2015batch} and LeakyReLu activations with a $0.2$ negative slope coefficient.

The proposed resizer model is relatively lightweight and does not add a significant number of trainable parameters to the baseline task. The number of trainable parameters for various configurations of the CNN are shown in Table~\ref{tab:cnn_params}. These CNNs are significantly smaller than a baseline model such as ResNet-50~\cite{he2016deep} which has about $23$ million parameters. Performance of these configurations are compared in Section~\ref{sec:ablation}, where we show that even the lightest configuration with $n=16,r=1$ is quite effective.

\subsection{Learning Losses}

\begin{table*}[t]
\vspace{-4mm}
\renewcommand{\arraystretch}{0.8}
\footnotesize
    \centering
    \begin{tabular}{lccccc} \toprule
        \multirow{2}{*} &  & Resizer & Baseline & \\
        Task &  Benchmark Data & Initialization & Initialization & Baseline Models & Training Loss\\\midrule
        Classification &  ImageNet~\cite{russakovsky2015imagenet} & Random & Pre-trained  & \makecell{Inception-v2~\cite{szegedy2016rethinking}, DenseNet-121~\cite{huang2017densely},\\ ResNet-50~\cite{he2016deep}, MobileNet-v2~\cite{sandler2018mobilenetv2}} & Cross-entropy \\\midrule
        IQA &  AVA~\cite{murray2012ava}  & Random & Pre-trained & \makecell{EfficientNet-b0~\cite{tan2019efficientnet}, Inception-v2~\cite{szegedy2016rethinking}, \\ DenseNet-121~\cite{huang2017densely}} & EMD \\\midrule
    \end{tabular}
    \vspace{-0.5mm}
    \caption{Summary of the tasks implemented in this paper. The learned resizer shown in Figure~\ref{fig:cnn} is jointly trained with each baseline model.}   
    \label{tab:task_summary}
\vspace{-2mm}
\end{table*}

The resizer is jointly trained with the baseline model loss. Since our objective is to learn an optimal resizer for a baseline vision task, we do not apply any loss or regularization constraint on the resized image. A summary of the tasks explored in this paper is shown in Table~\ref{tab:task_summary}. 

\subsubsection{Image Classification}

The classification models are trained with the cross-entropy loss. More specifically, the loss is computed on the final logits with a Sigmoid layer. The ImageNet~\cite{russakovsky2015imagenet} classification challenge consists of 1000 object classes, hence, the final logits layer represents 1000 predicted classes. We also use the label-smoothing regularization proposed by Szegedy et al.~\cite{szegedy2016rethinking}. The recognition loss can be expressed as
\begin{align}
\label{eq:cross_entropy}
L_\text{recog} = -\sum_{k=1}^{K} \mbox{log}(p_k)q'_k
\end{align}
\noindent where $p$ and $q'$ are the predictions and the smoothed labels, respectively, and $K$ denotes the total number of classes. The smoothed label of an image with a ground truth label $y$ is computed as $q'_k = (1-\epsilon)\delta_{k,y} + \epsilon/K$, where $\delta_{k,y}$ is $1$ when $k=y$, and $0$ otherwise. We keep $\epsilon$ fixed as $0.1$. The label regularization prevents the largest logit from dominating the other logits, leading to a less confident model and less over-fitting. 

\subsubsection{Image Quality Assessment (IQA)}

Our quality assessment models are trained through regression loss. Each image in the AVA dataset~\cite{murray2012ava} has a histogram of human ratings, with scores ranging from 1 to 10. Following the recent work in~\cite{talebi2018nima}, we use the Earth Mover's Distance (EMD) as our training loss. More specifically, the last layer of the baseline model is modified to have 10 logits, with a Softmax layer. The EMD loss can be represented as
\begin{align}
\label{eqn:emd}
L_{quality} = \left( \frac{1}{K} \sum_{k=1}^{K} |\mbox{CDF}(p_k) - \mbox{CDF}(q_k)|^d \right)^{1/d}
\end{align}
\noindent where $\mbox{CDF}(.)$ is the cumulative distribution function. In our implementation we found $d=2$ to be the most effective. Also, note that $K$ is equal to $10$ for the AVA dataset~\cite{murray2012ava}. The EMD loss accommodates learning the distribution of human ratings. This has proven to be more effective than regressing to the mean ratings.

\section{Experiments}
\label{sec:experiments}

Table~\ref{tab:task_summary} shows a summary of the specifics of our experiments. First, we train the baseline models on each dataset without the proposed resizer. For these cases we use the bilinear and the bicubic methods. These models are used as benchmarks to measure performance of the learned resizer. We also use these baselines to initialize the classification and IQA CNNs. For each baseline model and task, a separate resizer CNN is jointly trained with the baseline model. The resizer weights are randomly initialized. 

To showcase the impact of the proposed resizer, we train the baseline model at various image resolutions with and without the resizer (shown in Table~\ref{tab:classification_models} and Table~\ref{tab:quality_models}). More specifically, since the baseline models can be trained at any resolution, we vary the input size from the default $224\times224$ size to a larger $448\times448$ resolution. We use mini-batch learning, therefore the input image dimensions must be equal in each batch. To achieve this, and also allow the proposed resizer to train at higher resolutions, images in ImageNet and AVA are first resized by the bilinear or the bicubic method to a fixed resolution. The resizer's input resolution is always kept greater than or equal to its output resolution. Also, we apply the same resizing configuration at train and test time. 

Feeding higher resolution images to CNNs leads to higher usage of computational resources. In principle, this extra computation should be justified by a corresponding boost in performance. This applies to all our experimental models with and without the learned resizer. To provide a fair comparison from the computational standpoint, in our experiments the floating point operations per second (FLOPS) are also reported (shown in Table~\ref{tab:classification_models} and Table~\ref{tab:quality_models}). 

We use Tensorflow~\cite{abadi2016tensorflow} to train our networks with stochastic gradient descent with 4
NVIDIA V100 GPUs. Throughout our experiments we used the momentum optimizer~\cite{sutskever2013importance} with a decay of $0.9$. We used a learning rate of $0.05$ when training from scratch, and $0.005$ for fine-tuning. The learning rate is decayed every two epochs using an exponential rate of $0.94$. Next, we discuss our results. 

\begin{table*}[t]
\vspace{-0mm}
\renewcommand{\arraystretch}{0.83}
\footnotesize
    \centering
    \begin{tabular}{lccccccc} \toprule
        \makecell{Classification \\ Model} & Resizer & \makecell{Resizer's Input \\ Resolution} & \makecell{Resizer's Output \\ Resolution} & \makecell{Batch \\ Size} & Top-1 Error $\downarrow$ & Top-5 Error $\downarrow$ & \makecell{Total FLOPS \\ (Billions)} \\\midrule
        \parbox[c]{5mm}{\multirow{9}{*}{\rotatebox[origin=c]{90}{\normalsize{Inception-v2}~\cite{szegedy2016rethinking}}}}
        & Proposed  & 224$\times$224 &	224$\times$224 & 128 & 24.1\% & 7.5\% & 5.07\\
        & Proposed  & 256$\times$256 &	224$\times$224 & 96 & \textbf{24.0}\% & 7.4\% & 5.15 \\
        & Proposed  & 320$\times$320 &	224$\times$224 & 64 & 24.2\% & \textbf{7.3}\% & 5.35 \\
        & Proposed  & 368$\times$368 &	224$\times$224 & 48 & 24.5\% & 7.4\% & 5.52 \\
        & Proposed  & 448$\times$448 &	224$\times$224 & 32 & 25.3\% & 7.9\% & 5.86 \\\cdashline{2-8}
        & Bilinear  & original &	224$\times$224 & 128 & 26.7\% & 8.7\% & 3.88 \\\cdashline{2-8}
        & Bilinear  & original &	256$\times$256 & 96 & 27.3\% & 9.1\% & 5.07 \\
        & Bilinear  & original &	320$\times$320 & 64 & 27.3\% & 9.1\% & 7.92 \\
        & Bilinear  & original &	368$\times$368 & 48 & 29.6\% & 10.4\% & 10.6 \\
        & Bilinear  & original &	448$\times$448 & 32 & 30.6\% & 11.2\% & 15.52 \\\midrule
        \parbox[c]{0mm}{\multirow{9}{*}{\rotatebox[origin=c]{90}{\normalsize{DenseNet-121}~\cite{huang2017densely}}}}
        & Proposed  & 224$\times$224 &	224$\times$224 & 128 & 31.1\% & 11.6\% & 6.86\\
        & Proposed  & 256$\times$256 &	224$\times$224 & 96 & 31.0\% & 11.4\% & 6.95 \\
        & Proposed  & 320$\times$320 &	224$\times$224 & 64 & 30.7\% & 11.1\% & 7.14 \\
        & Proposed  & 368$\times$368 &	224$\times$224 & 48 & 30.2\% & 10.9\% & 7.31 \\
        & Proposed  & 448$\times$448 &	224$\times$224 & 32 & \textbf{29.8}\% & \textbf{10.8}\% & 7.65 \\\cdashline{2-8}
        & Bilinear  & original &	224$\times$224 & 128 & 33.1\% & 12.8\% & 5.67 \\\cdashline{2-8}
        & Bilinear  & original &	256$\times$256 & 96 & 30.9\% & 11.7\% & 7.41 \\
        & Bilinear  & original &	320$\times$320 & 64 & 29.9\% & 10.8\% & 11.57 \\
        & Bilinear  & original &	368$\times$368 & 48 & 29.7\% & 10.7\% & 15.26 \\
        & Bilinear  & original &	448$\times$448 & 32 & 31.5\% & 12.0\% & 22.68 \\\midrule
        \parbox[c]{0mm}{\multirow{9}{*}{\rotatebox[origin=c]{90}{\normalsize{ResNet-50}~\cite{he2016deep}}}}
        & Proposed  & 224$\times$224 &	224$\times$224 & 128 & 23.7\% & 7.0\% & 8.16\\
        & Proposed  & 256$\times$256 &	224$\times$224 & 96 & 23.8\% & 7.0\% & 8.24 \\
        & Proposed  & 320$\times$320 &	224$\times$224 & 64 & 23.4\% & 6.8\% & 8.43 \\
        & Proposed  & 368$\times$368 &	224$\times$224 & 48 & \textbf{23.0}\% & \textbf{6.7}\% & 8.61 \\
        & Proposed  & 448$\times$448 &	224$\times$224 & 32 & 23.7\% & 6.9\% & 8.95 \\\cdashline{2-8}
        & Bilinear  & original &	224$\times$224 & 128 & 24.7\% & 7.5\% & 6.97 \\\cdashline{2-8}
        & Bilinear  & original &	256$\times$256 & 96 & 23.5\% & 6.9\% & 9.10 \\
        & Bilinear  & original &	320$\times$320 & 64 & 22.5\% & 6.3\% & 14.21 \\
        & Bilinear  & original &	368$\times$368 & 48 & 22.1\% & 6.0\% & 19.17 \\
        & Bilinear  & original &	448$\times$448 & 32 & 21.9\% & 5.8\% & 27.85 \\\midrule
        \parbox[c]{0mm}{\multirow{9}{*}{\rotatebox[origin=c]{90}{\normalsize{MobileNet-v2}~\cite{sandler2018mobilenetv2}}}}
        & Proposed  & 224$\times$224 &	224$\times$224 & 128 & 29.1\% & 10.1\% & 1.79\\
        & Proposed  & 256$\times$256 &	224$\times$224 & 96 & 29.0\% & 10.1\% &  1.87 \\
        & Proposed  & 320$\times$320 &	224$\times$224 & 64 & 28.7\% & 9.9\% & 2.07 \\
        & Proposed  & 368$\times$368 &	224$\times$224 & 48 & \textbf{28.4}\% & \textbf{9.8}\% & 2.24 \\
        & Proposed  & 448$\times$448 &	224$\times$224 & 32 & 28.5\% & \textbf{9.8}\% & 2.58 \\\cdashline{2-8}
        & Bilinear  & original &	224$\times$224 & 128 & 29.5\% & 10.4\% & 0.60 \\\cdashline{2-8}
        & Bilinear  & original &	256$\times$256 & 96 & 28.7\% & 9.6\% & 0.78 \\
        & Bilinear  & original &	320$\times$320 & 64 & 27.2\% & 9.0\% & 1.23 \\
        & Bilinear  & original &	368$\times$368 & 48 & 26.6\% & 8.6\% & 1.66 \\
        & Bilinear  & original &	448$\times$448 & 32 & 26.1\% & 8.3\% & 2.40 \\\midrule
    \end{tabular}
    \vspace{-0.5mm}
    \caption{Classification errors on the ImageNet~\cite{russakovsky2015imagenet} validation set using various models. Each row represents a model trained with a different resizing configuration. The highlighted results represent the best performances among all models with $224\times224$ input resolution. Note that as the input resolution increases, the batch size is reduced to avoid memory consumption issues. Also, images are resized to fix resolutions before feeding them to the proposed resizer (shown under resizer's input resolution) to accommodate for mini-batch gradient descent.}   
    \label{tab:classification_models}
    \vspace{-5mm}
\end{table*}

\begin{figure*}
    \centering
    \scriptsize

    \vspace{-4mm}

    \begin{overpic}[width=.27\linewidth]{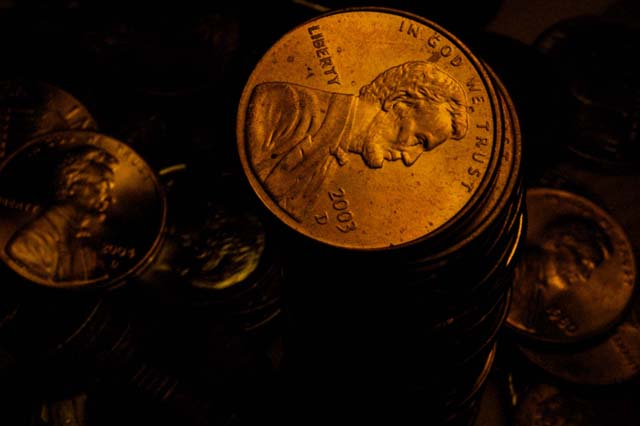}
    \put(25,2){\begin{color}{white}Original ($427\times640$)\end{color}}
    \end{overpic}
    \begin{overpic}[width=.24\linewidth]{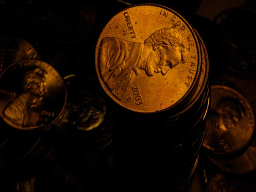}
    \put(20,2){\begin{color}{white} Bilinear ($192\times256$)\end{color}}
    \end{overpic}
    \begin{overpic}[width=.24\linewidth]{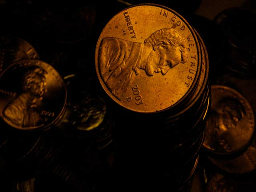}
    \put(20,2){\begin{color}{white} Bicubic ($192\times256$)\end{color}}
    \end{overpic}
    %
    %
    %
    %
    %
    %
    \begin{overpic}[width=.24\linewidth]{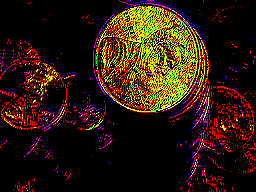}
    \put(16,2){\begin{color}{white} Inception-v2 ($192\times256$) \end{color}}
    \end{overpic}
    \begin{overpic}[width=.24\linewidth]{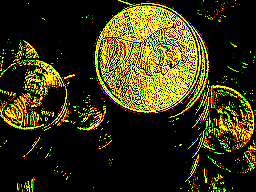}
    \put(16,2){\begin{color}{white}DenseNet-121 ($192\times256$) \end{color}}
    \end{overpic}
    \begin{overpic}[width=.24\linewidth]{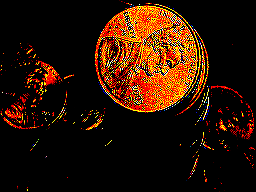}
    \put(18,2){\begin{color}{white}ResNet-50 ($192\times256$) \end{color}}
    \end{overpic}
    \begin{overpic}[width=.24\linewidth]{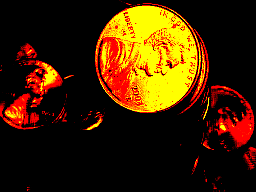}
    \put(16,2){\begin{color}{white}MobileNet-v2 ($192\times256$) \end{color}}
    \end{overpic}
    
    \vspace{-1mm}
    
    \caption{Examples of the proposed learned resizer trained together with various classification models on ImageNet~\cite{russakovsky2015imagenet}. The resizers lead to improved recognition performances.}
    \label{fig:classification_resized1}
\end{figure*}

\begin{figure*}
    \centering
    \scriptsize

    \vspace{-3mm}
    \begin{overpic}[width=.145\linewidth]{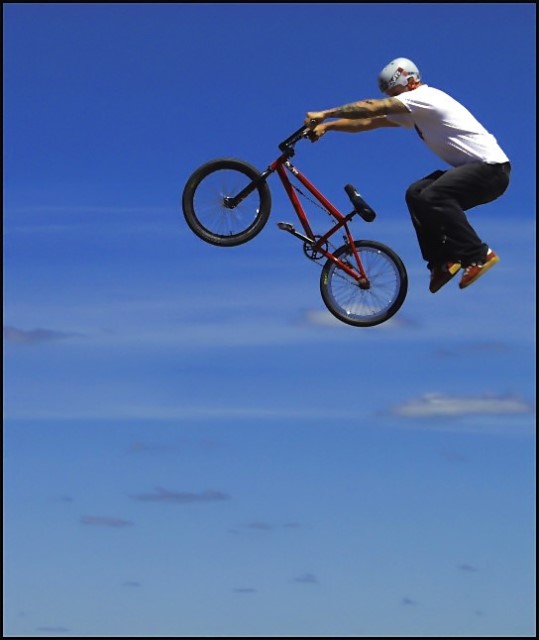}
    \put(2,90){\begin{color}{white}Original \end{color}}
    \put(2,2){\begin{color}{white} ($640\times539$)\end{color}}
    \end{overpic}
    \begin{overpic}[width=.13\linewidth]{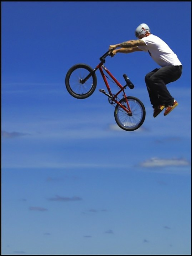}
    \put(2,90){\begin{color}{white} Bilinear \end{color}}
    \put(2,2){\begin{color}{white} ($256\times192$)\end{color}}
    \end{overpic}
    \begin{overpic}[width=.13\linewidth]{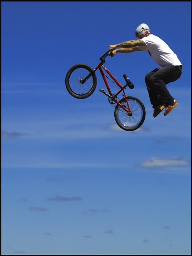}
    \put(2,90){\begin{color}{white} Bicubic \end{color}}
    \put(2,2){\begin{color}{white} ($256\times192$)\end{color}}
    \end{overpic}
    \begin{overpic}[width=.13\linewidth]{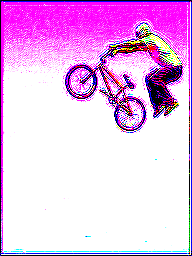}
    \put(2,90){\begin{color}{black} Inception-v2 \end{color}}
    \put(2,2){\begin{color}{black} ($256\times192$)\end{color}}
    \end{overpic}
    \begin{overpic}[width=.13\linewidth]{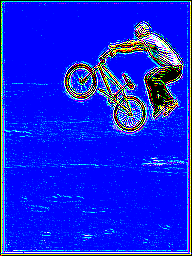}
    \put(2,90){\begin{color}{white}DenseNet-121 \end{color}}
    \put(2,2){\begin{color}{white} ($256\times192$)\end{color}}
    \end{overpic}
    \begin{overpic}[width=.13\linewidth]{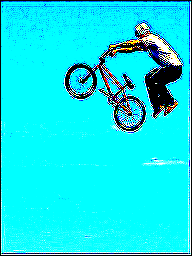}
    \put(2,90){\begin{color}{black}ResNet-50\end{color}}
    \put(2,2){\begin{color}{black} ($256\times192$)\end{color}}
    \end{overpic}
    \begin{overpic}[width=.13\linewidth]{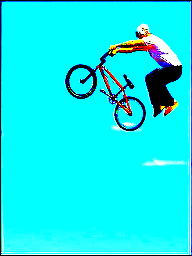}
    \put(2,90){\begin{color}{black}MobileNet-v2 \end{color}}
    \put(2,2){\begin{color}{black} ($256\times192$)\end{color}}
    \end{overpic}
    
    \vspace{-1mm}
    
    \caption{Examples of the proposed learned resizer trained with various classification models on ImageNet~\cite{russakovsky2015imagenet}. The resizers lead to improved recognition performances.}
    \label{fig:classification_resized2}
    \vspace{-5mm}
\end{figure*}

\begin{table*}[t]
\vspace{-3mm}
\renewcommand{\arraystretch}{0.9}
\footnotesize
    \centering
    \begin{tabular}{lccccccc} \toprule
        \makecell{Baseline \\ Model} & Resizer & \makecell{Resizer's Input \\ Resolution} & \makecell{Resizer's Output \\Resolution} & \makecell{Batch \\ Size} & PLCC $\uparrow$ & SRCC $\uparrow$ & \makecell{Total FLOPS \\ (Billions)} \\\midrule
        \parbox[c]{0mm}{\multirow{9}{*}{\rotatebox[origin=c]{90}{\normalsize{Inception-v2}~\cite{szegedy2016rethinking}}}}
        & Proposed  & 224$\times$224 &	224$\times$224 & 128 & 0.673 & 0.653 & 5.07 \\
        & Proposed  & 256$\times$256 &	224$\times$224 & 96 & 0.674 & 0.655  & 5.15 \\
        & Proposed  & 320$\times$320 &	224$\times$224 & 64 & \textbf{0.686} & \textbf{0.663} & 5.35 \\
        & Proposed  & 368$\times$368 &	224$\times$224 & 48 & 0.677 & 0.652  & 5.52 \\
        & Proposed  & 448$\times$448 &	224$\times$224 & 32 & 0.677 & 0.651  & 5.86 \\\cdashline{2-8}
        & Bicubic  & original &	224$\times$224 & 128 & 0.662 & 0.643 & 3.88 \\\cdashline{2-8}
        & Bicubic  & original &	256$\times$256 & 96 & 0.672 & 0.652  & 5.07 \\
        & Bicubic  & original &	320$\times$320 & 64 & 0.688 & 0.664 & 7.92 \\
        & Bicubic  & original &	368$\times$368 & 48 & 0.693 &  0.668 & 10.6 \\
        & Bicubic  & original &	448$\times$448 & 32 & 0.700 & 0.672  & 15.52 \\\midrule
        \parbox[c]{0mm}{\multirow{9}{*}{\rotatebox[origin=c]{90}{\normalsize{DenseNet-121}~\cite{huang2017densely}}}}
        & Proposed  & 224$\times$224 &	224$\times$224 & 128 & 0.672 & 0.644  & 6.86 \\
        & Proposed  & 256$\times$256 &	224$\times$224 & 96 & 0.672 &  0.645 & 6.95 \\
        & Proposed  & 320$\times$320 &	224$\times$224 & 64 & \textbf{0.683} & \textbf{0.655} & 7.14 \\
        & Proposed  & 368$\times$368 &	224$\times$224 & 48 & 0.675 & 0.644 & 7.31 \\
        & Proposed  & 448$\times$448 &	224$\times$224 & 32 & 0.673 & 0.642 & 7.65 \\\cdashline{2-8}
        & Bicubic  & original &	224$\times$224 & 128 & 0.662 & 0.636  & 5.67 \\\cdashline{2-8}
        & Bicubic  & original &	256$\times$256 & 96 & 0.672 &  0.644 & 7.41 \\
        & Bicubic  & original &	320$\times$320 & 64 & 0.694 & 0.666 & 11.57 \\
        & Bicubic  & original &	368$\times$368 & 48 & 0.695 & 0.663 & 15.26 \\
        & Bicubic  & original &	448$\times$448 & 32 & 0.692 & 0.658 & 22.68 \\\midrule
        \parbox[c]{0mm}{\multirow{9}{*}{\rotatebox[origin=c]{90}{\normalsize{EfficientNet-b0}~\cite{tan2019efficientnet}}}}
        & Proposed  & 224$\times$224 &	224$\times$224 & 128 & 0.646 & 0.626  & 1.93 \\
        & Proposed  & 256$\times$256 &	224$\times$224 & 96 & 0.650 & 0.629  & 2.01 \\
        & Proposed  & 320$\times$320 &	224$\times$224 & 64 & \textbf{0.671} & \textbf{0.651}  & 2.20 \\
        & Proposed  & 368$\times$368 &	224$\times$224 & 48 & 0.654 & 0.632 & 2.38 \\
        & Proposed  & 448$\times$448 &	224$\times$224 & 32 & 0.644 & 0.616  & 2.72 \\\cdashline{2-8}
        & Bicubic  & original &	224$\times$224 & 128 & 0.642 & 0.620  & 0.74 \\\cdashline{2-8}
        & Bicubic  & original &	256$\times$256 & 96 & 0.659 & 0.637 & 0.97 \\
        & Bicubic  & original &	320$\times$320 & 64 & 0.674 & 0.652 & 1.51 \\
        & Bicubic  & original &	368$\times$368 & 48 & 0.678 & 0.655 & 2.05 \\
        & Bicubic  & original &	448$\times$448 & 32 & 0.673 & 0.648  & 2.96 \\\midrule
    \end{tabular}
    \vspace{-0.5mm}
    \caption{IQA on the AVA dataset~\cite{murray2012ava} with various models. Each row represents a model trained with a different resizing configuration. Performance of each model is quantified by the Pearson and Spearman correlations of the predicted and ground truth mean scores. The highlighted results represent the best performances among all models with $224\times224$ input resolution. Note that as the input resolution increases, the batch size is reduced to avoid memory consumption issues. Also, images are resized to fix resolutions before feeding them to the proposed resizer (shown under resizer's input resolution) to accommodate for mini-batch gradient descent.}   
    \label{tab:quality_models}
\end{table*}

\subsection{Classification}
\label{sec:experimental_classification}

We select four baseline models to jointly train with the image resizer. We present our results for the ImageNet dataset in Table~\ref{tab:classification_models}. We experiment with various resolutions and adjust the batch size to avoid exceeding our memory limits. The reported top-k error is the percentage of time that the classifier does not return the correct class in the top $k$ highest probability scores. We call the model trained with bilinear resizer and output resizing resolution $224\times224$ the default baseline. The highlighted results represent the best performances among models with $224\times224$ resolution. As can be seen, networks trained with the proposed resizer show an overall improvement over the default baseline. Comparing to the default baseline, DenseNet-121 and MobileNet-v2 baselines show the largest and smallest gains, respectively. Also, it is worth mentioning that for the Inception-v2, DenseNet-121, and ResNet-50 models, the proposed resizer performs better than the bilinear resizer with comparable FLOPS. However, training the MobileNet-v2 model with bilinear resizer at higher resolutions is more effective than using the learned resizer with similar FLOPS.

Table~\ref{tab:classification_models} also shows that with or without the proposed resizer, increasing the input resolution benefits the performance of DenseNet-121, ResNet-50, and MobileNet-v2. The Inception-v2 model is an exception as it gains the most performance boost from training with larger batch sizes. It is worth noting that training resizers with equal input and output resolutions also results in improvement over the default baselines. In most cases, however, the best performance is obtained when the resizer's input is larger than its output.

We also present some examples to visually compare the trained resizers in Figure~\ref{fig:classification_resized1} and Figure~\ref{fig:classification_resized2}. Perhaps the common feature among these results is the boost of the high frequency details. Interestingly these effects tend to make the classification model more effective. Aside from the MobileNet-v2 results, the other models tend to create overly sharpened results. This may intuitively explain the low performance gain obtained for MobileNet-v2. Overall, these effects do not meet the perceptual bar for human vision, but they surely improve the machine vision task.

\subsection{Quality Assessment}
\label{sec:experimental_quality}

We use 3 different baseline models to train with the AVA dataset~\cite{murray2012ava}. The baseline models are initialized from pre-trained weights on ImageNet~\cite{russakovsky2015imagenet}, and fine-tuned on the AVA dataset. Note that the resizer weights are initialized randomly. In this set of experiments we use the bicubic resizer as our baseline method. Our results are presented in Table~\ref{tab:quality_models}. We measure the performance by reporting correlation between the mean ground truth score and mean predicted score. To this end, we use the Pearson linear correlation coefficient (PLCC), and Spearman rank correlation coefficient (SRCC). As can be seen, there is a consistent improvement over the baseline models. Also, it is worth noting that for Inception-v2 and DenseNet-121 models, the proposed resizer performs better than the bicubic resizer with comparable FLOPS. At higher FLOPS, EfficientNet seems to be a more challenging baseline for the learned resizer.

\begin{figure*}
    \centering
    \scriptsize

    \vspace{-3mm}
    \begin{overpic}[width=.245\linewidth]{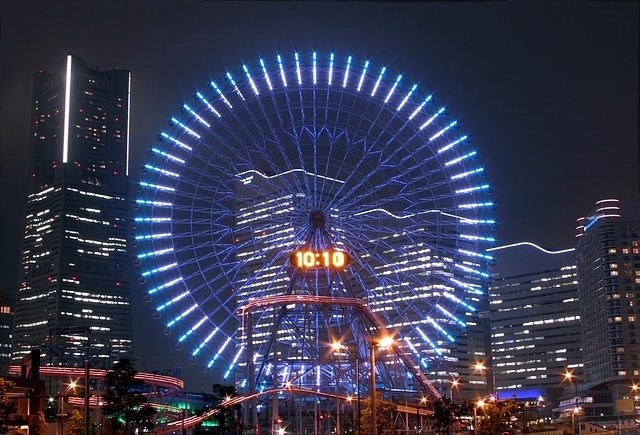}
    \put(2,2){\begin{color}{white} (a) Original ($435\times640$)\end{color}}
    \end{overpic}
    \begin{overpic}[width=.245\linewidth]{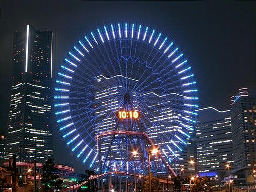}
    \put(2,2){\begin{color}{white} (c) Inception-v2 ($192\times256$)\end{color}}
    \end{overpic}
    \begin{overpic}[width=.245\linewidth]{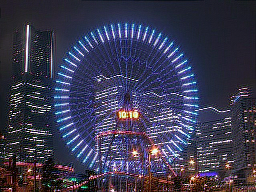}
    \put(2,2){\begin{color}{white} (e) DenseNet-121 ($192\times256$)\end{color}}
    \end{overpic}
    \begin{overpic}[width=.245\linewidth]{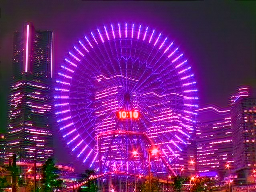}
    \put(2,2){\begin{color}{white} (g) EfficientNet-b0 ($192\times256$)\end{color}}
    \end{overpic}
    \begin{overpic}[width=.245\linewidth]{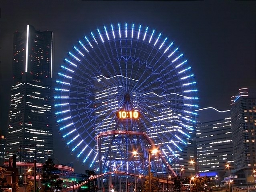}
    \put(2,2){\begin{color}{white} (b) Bicubic ($192\times256$) \end{color}}
    \end{overpic}
    \begin{overpic}[width=.245\linewidth]{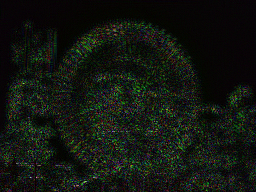}
    \put(2,2){\begin{color}{white} (d) $|$c - b$|$ ($192\times256$)\end{color}}
    \end{overpic}
    \begin{overpic}[width=.245\linewidth]{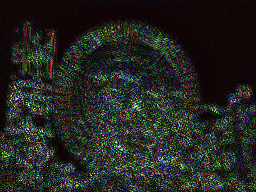}
    \put(2,2){\begin{color}{white} (f) $|$e - b$|$ ($192\times256$)\end{color}}
    \end{overpic}
    \begin{overpic}[width=.245\linewidth]{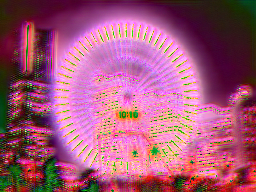}
    \put(2,2){\begin{color}{white} (h) $|$g - b$|$ ($192\times256$)\end{color}}
    \end{overpic}
    \vspace{-1mm}
    
    \caption{Examples of the proposed learned resizer trained with various IQA models on the AVA dataset~\cite{murray2012ava}. (c), (e), and (f) are results from trained resizers with respective base models. (d), (f), and (h) represent the difference between bicubic and learned resizers.}
    \label{fig:quality_resized1}
    \vspace{-5mm}
\end{figure*}

Examples of the trained resizers are shown in Figure~\ref{fig:quality_resized1}. The residual images show the difference between the bicubic and the learned resizer. As can be seen, the residual image for the Inception and the DenseNet models represent a lot of fine grain details. On the other hand, the EfficientNet resizer shows a strong color shift and modest detail manipulations.



\subsection{Generalization}

In this section generalization of the resizer model is discussed. To this end, we first jointly fine-tune the learned resizer with a target baseline that is different from the resizer's default baseline. Then, we measure performance of the target baseline on the underlying task. We observed that fine-tuning on about 4 epochs of the training data suffices to adapt the resizer to the target model. This validation is a reasonable indicator of how well the trained resizers generalize to various architectures. Our results for classification and IQA are presented in Table~\ref{tab:classification_generalization} and Table~\ref{tab:quality_generalization}. Each column shows an initialization checkpoint for the resizer model, and each row indicates a target baseline. These results show that a resizer trained for one baseline can be effectively used to develop a resizer for another baseline with minimal fine-tuning. In some cases such as the DenseNet and MobileNet models the fine-tuned resizers actually surpass the classification performance obtained by random initialization (see Table~\ref{tab:classification_generalization}). The same observation holds true for the EfficientNet model in the IQA application (see Table~\ref{tab:quality_generalization}). These improvements are perhaps because of the transfer learning effect.

We also tried the above cross-model validation without fine-tuning, however, performance mostly degraded. This is likely because (1) the proposed resizer is exclusively trained for one baseline model, and (2) no intermediate pixel loss is used during training.

For cross-dataset validation we train the resizer on ImageNet, and test it with a classifier trained on the CIFAR-10 benchmark~\cite{krizhevsky2009learning}. With the addition of the learned resizer to the classification task on CIFAR-10, the top-1 error of ResNet-50 (6.9\%) baseline drops by 0.4\%. We observed a similar favorable trend in other baseline models.



\begin{table*}[t]
\vspace{-2mm}
\renewcommand{\arraystretch}{0.9}
\footnotesize
    \centering
    \begin{tabular}{lcccccccccc} \toprule
        \multirow{2}{*} &  \multicolumn{4}{c}{Top-1 Error $\downarrow$}  &  &  \multicolumn{4}{c}{Top-5 Error $\downarrow$} \\ \cmidrule{2-5} \cmidrule{7-10}
        \diagbox[width=7em]{Target}{Initial} & Inception-v2 & DenseNet-121 &  ResNet-50 &  MobileNet-v2 & & Inception-v2 & DenseNet-121 &  ResNet-50 &  MobileNet-v2 \\\cmidrule{1-10}
        Inception-v2 & 24.5\%  & 24.6\% & 24.5\%  & 24.6\% & & 7.4\%  & 7.5\% & 7.5\%  & 7.4\%  \\
        DenseNet-121 & 29.7\%  & 30.2\% & 29.7\%  & 30.1\% & & 10.7\%  & 10.9\% & 10.6\%  & 10.9\%  \\
        ResNet-50 & 22.9\%  & 23.0\% & 23.0\%  & 23.1\% & & 6.5\%  & 6.4\% & 6.7\%  & 6.7\%  \\
        MobileNet-v2 & 28.0\%  & 28.2\% & 28.0\%  & 28.4\% & & 9.6\%  & 9.6\% & 9.6\%  & 9.8\%  \\\midrule
    \end{tabular}
    \vspace{-2mm}
    \caption{Generalization of the resizer models for image classification~\cite{russakovsky2015imagenet}. The learned resizers are trained with the initial baseline and then jointly fine-tuned with the target baseline model. The resizer's input and output resolutions are $368\times368$ and $224\times224$, respectively.}
    \label{tab:classification_generalization}
    \vspace{-2mm}
\end{table*}

\begin{table*}[t]
\renewcommand{\arraystretch}{0.6}
\footnotesize
    \centering
    \begin{tabular}{lcccccccc} \toprule
        \multirow{2}{*} &  \multicolumn{3}{c}{PLCC $\uparrow$}  &  &  \multicolumn{3}{c}{SRCC $\uparrow$} \\ \cmidrule{2-4} \cmidrule{6-8}
        \diagbox[width=7em]{Target}{Initial} & Inception-v2& DenseNet-121& EfficientNet-b0 & & Inception-v2 & DenseNet-121 &  EfficientNet-b0 &  \\\cmidrule{1-8}
        Inception-v2~\cite{szegedy2016rethinking} & 0.677 & 0.672  & 0.670 & & 0.652  & 0.649  & 0.649 \\
        DenseNet-121~\cite{huang2017densely} & 0.672  & 0.675  & 0.671 & & 0.645  & 0.644  & 0.642 \\
        EfficientNet-b0~\cite{tan2019efficientnet} & 0.660 & 0.653 & 0.654 & & 0.636 & 0.632 & 0.632 \\\midrule
    \end{tabular}
    \vspace{-0.5mm}
    \caption{Generalization of the resizer models for IQA~\cite{murray2012ava}. The learned resizers are trained with the initial baseline and then jointly fine-tuned with the target baseline model. The resizer's input and output resolutions are $368\times368$ and  $224\times224$, respectively.}
    \label{tab:quality_generalization}
    \vspace{-2mm}
\end{table*}

\begin{table*}[!t]
\renewcommand{\arraystretch}{0.6}
\footnotesize
    \centering
    \begin{tabular}{lccccccccccc} \toprule
        \multirow{2}{*} & &  \multicolumn{4}{c}{Top-1 Error $\downarrow$}  &  &  \multicolumn{4}{c}{Top-5 Error $\downarrow$} \\ \cmidrule{3-6} \cmidrule{8-11}
        \makecell{Task} & Model & \makecell{$r=1$ \\ $n=16$} & \makecell{$r=2$ \\ $n=16$} &  \makecell{$r=1$ \\ $n=32$} &  \makecell{$r=2$ \\ $n=32$} & & \makecell{$r=1$ \\ $n=16$} & \makecell{$r=2$ \\ $n=16$} &  \makecell{$r=1$ \\ $n=32$} &  \makecell{$r=2$ \\ $n=32$} \\\cmidrule{1-11}
        \parbox[c]{0mm}{\multirow{4}{*}{\rotatebox[origin=c]{90}{\scriptsize{Classification}}}}
        & Inception-v2~\cite{szegedy2016rethinking} & 24.5\%  & 25.5\% & 25.6\%  & 26.1\% & & 7.4\%  & 8.0\% & 7.9\%  & 8.3\%  \\
        & DenseNet-121~\cite{huang2017densely} & 30.2\%  & 29.8\% & 29.9\%  & 29.8\% & & 10.9\%  & 10.8\% & 10.8\%  & 10.8\%  \\
        & ResNet-50~\cite{he2016deep} & 23.0\%  & 23.4\% & 23.4\%  & 23.3\% & & 6.7\%  & 6.6\% & 6.7\%  & 6.8\%  \\
        & MobileNet-v2~\cite{sandler2018mobilenetv2} & 28.4\%  & 28.5\% & 28.4\%  & 28.3\% & & 9.8\%  & 9.7\% & 9.7\%  & 9.7\%  \\ 
        \cmidrule{3-11} 
        \multirow{2}{*} & &  \multicolumn{4}{c}{PLCC $\uparrow$}  &  &  \multicolumn{4}{c}{SRCC $\uparrow$} \\ \cmidrule{3-6} \cmidrule{8-11} 
        \parbox[c]{0mm}{\multirow{3}{*}{\rotatebox[origin=c]{90}{\scriptsize{IQA}}}}
        & Inception-v2~\cite{szegedy2016rethinking} & 0.677  & 0.677 & 0.675  & 0.676 &  & 0.652 & 0.654 & 0.643 & 0.643 \\
        & DenseNet-121~\cite{huang2017densely} & 0.675  & 0.677 & 0.670  & 0.671 &  & 0.644 & 0.645 & 0.629 & 0.630 \\
        & EfficientNet-b0~\cite{tan2019efficientnet} & 0.654  & 0.652 & 0.646  & 0.648 &  & 0.632 & 0.630 & 0.625 & 0.628 \\\midrule
    \end{tabular}
    \vspace{-0.5mm}
    \caption{Effect of the resizer model parameters on the classification~\cite{russakovsky2015imagenet}, and image quality assessment (IQA)~\cite{murray2012ava}. Parameters $r$ and $n$ denote the number of residual blocks and convolutional filters, respectively. These parameters are presented in Figure~\ref{fig:cnn}. The resizer's input and output resolutions are $368\times368$ and $224\times224$, respectively.}
    \label{tab:ablation}
    \vspace{-3mm}
\end{table*}

\begin{table}[h!]
\vspace{-1mm}
\begin{center}
\footnotesize
\renewcommand{\arraystretch}{0.55}
\setlength{\tabcolsep}{3.5pt}
\begin{tabular}{rcccc} 
Model & Pre-processor & Down-scaled (4$\times$) & Noise & JPEG \\\cmidrule{1-5}
\multirow{ 2}{*}{ResNet-50} & \cite{liu2019transferable}    & 31.8\%	& 29.1\% & 34.9\% \\
 & ours  & 31.1\% & 28.6\%  & 34.5\%  \\\cmidrule{1-5}
\end{tabular}
\end{center}
\vspace{-4mm}
\caption{Top-1 classification error on distorted and then enhanced ImageNet images. Results from \cite{liu2019transferable} are their best performing models.} \vspace{-5mm}
\label{tab:comp}
\end{table}

We also compare our performance with \cite{liu2019transferable} in Table~\ref{tab:comp}. We followed the authors instructions in \cite{liu2019transferable} to generate the distorted images with Gaussian noise (standard deviation 0.1), spatial down-scaling (4$\times$), and JPEG compression with quality factor 10. We trained the resizer to increase the resolution by a factor of 4 for the down-scaled inputs, and kept the resolution unchanged for the other distortions. As can be noted from~Table~\ref{tab:comp}, the proposed resizer outperforms \cite{liu2019transferable} consistently. Also, it is worth pointing out that our pre-processor model obtains the largest margin of improvement for down-scaled images.

\subsection{Ablation}
\label{sec:ablation}

In this section effects of our design choices in the resizer model are discussed. We vary the number of residual blocks $r$, and the number of filters $n$ (see Figure~\ref{fig:cnn}), and report the performance of the jointly trained baseline models. Note that so far in the experimental results we have employed CNN resizers with the default configuration in which $r=1$ and $n=16$.

Our results for the classification and IQA tasks with various configurations are presented in Table~\ref{tab:ablation}. In the classification task, as the resizer model gets bigger, the DenseNet and the MobileNet baselines show modest improvements over the default configuration. However, Inception and ResNet do not benefit from larger number of parameters in the resizer. A similar trend can be observed in the IQA task.

Perhaps one of the reasons for the non-growing performance of the larger resizer models is the lowered batch size. Note that given limited memory, larger resizers have to be trained with smaller batch sizes. This factor may inadvertently limit the observed performance gain. 

\section{Conclusions}
\label{sec:conclusion}

We presented a framework for learning pre-processing effects that boost the performance of image recognition models. We focused on image resizing, and did not apply an intermediate pixel or perceptual loss on the reszied images, hence the results are exclusively optimized for machine vision tasks. Our experiments show that task-optimized deep vision models can benefit from replacing traditional image resizers with learned resizers. We believe that customized pre-processing algorithms for machine vision tasks have not been studied extensively, and given the impact shown in this paper, there is significant room for research in this area. As part of future work we will extend our model to other vision tasks. 

{\small
\bibliographystyle{ieee_fullname}
\bibliography{egbib}
}

\end{document}